%% file: main.tex
\documentclass[runningheads]{llncs}
\usepackage{hyperref}
\usepackage{amssymb}
\usepackage{booktabs}
\usepackage[T1]{fontenc}
\usepackage{booktabs}
\usepackage{multirow}
\usepackage[table,xcdraw]{xcolor}
\usepackage[frozencache,cachedir=.]{minted}

\usepackage{placeins}
\usepackage{listings}
\usepackage{enumitem}

\usepackage{todonotes}

\lstdefinestyle{myCustomMatlabStyle}{
  language=Python,
  stepnumber=1,
  numbersep=10pt,
  tabsize=4,
  showspaces=false,
  showstringspaces=false,
  frame=single,
}

\usepackage{graphicx}
\usepackage{cite}
\usepackage[table,xcdraw]{xcolor}
\usepackage{amsmath}
\usepackage{wrapfig}
\usepackage{gensymb}
\usepackage{rotating}
\usepackage[toc,page]{appendix}
\usepackage{bbding}

\usepackage{color}
  \definecolor{IMV1}{rgb}{0.64, 0.0, 0.0}
  \definecolor{IMV2}{rgb}{0.03, 0.27, 0.49}

\begin{document}
  
\title{Neural Network Verification for Gliding Drone Control: A Case Study}
\author{Colin Kessler\inst{1,2} \Envelope  \and
Ekaterina Komendantskaya \inst{1} \and
Marco Casadio\inst{1}\and
Ignazio Maria %remember that Maria is part of the first name and is not middle name 
Viola\inst{2} \and
Thomas Flinkow\inst{3} \and
Albaraa Ammar Othman\inst{1} \and
Alistair Malhotra\inst{1} \and
Robbie McPherson\inst{1}
}
\authorrunning{C. Kessler et al.}
\institute{Heriot-Watt University and
Edinburgh Centre for Robotics, UK\\ 
\email{ck2049@hw.ac.uk} \and
School of Engineering, University of Edinburgh, UK\and
Maynooth University, Maynooth, Ireland\\
}

\maketitle              % typeset the header of the contribution
\begin{abstract}
As machine learning is increasingly deployed in autonomous systems, verification of neural network controllers is becoming an active research domain. Existing tools and annual verification competitions suggest that soon this technology will become effective for real-world applications. Our application comes from the emerging field of microflyers that are passively transported by the wind, which may have various uses in weather or pollution monitoring. Specifically, we investigate centimetre-scale bio-inspired gliding drones that resemble \textit{Alsomitra macrocarpa} diaspores. In this paper, we propose a new case study on verifying \textit{Alsomitra}-inspired drones with neural network controllers, with the aim of adhering closely to a target trajectory. We show that our system differs substantially from existing VNN and ARCH competition benchmarks, and show that a combination of tools holds promise for verifying such systems in the future, if certain shortcomings can be overcome. We propose a novel method for robust training of regression networks, and investigate formalisations of this case study in Vehicle and CORA. Our verification results suggest that the investigated training methods do improve performance and robustness of neural network controllers in this application, but are limited in scope and usefulness. This is due to systematic limitations of both Vehicle and CORA, and the complexity of our system reducing the scale of reachability, which we investigate in detail. If these limitations can be overcome, it will enable engineers to develop safe and robust technologies that improve people’s lives and reduce our impact on the environment.

\end{abstract}
\section{Introduction}
\begin{figure*}[!h]
  \begin{center}
    \includegraphics[width=0.7\textwidth]{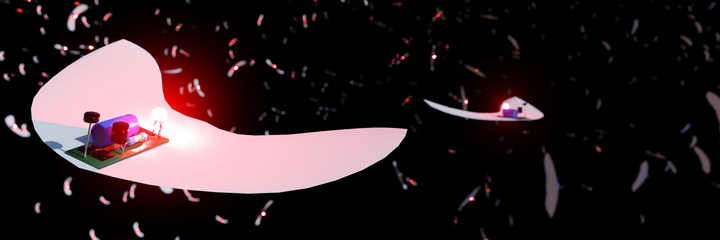}\vspace{-2mm}
  \end{center}
  \caption{An artist's impression of a swarm of gliding drones inspired by \textit{Alsomitra} seeds~\cite{certini2023alsomitra}.}\label{fig:also}
\end{figure*}

A recent research trend in drone design concerns the development of gliding microdrones, which could serve a function as airborne sensors and remain aloft for extended periods of time~\cite{electronic_microfliers, Iyer2022Wind, johnson2023origami, alsomitra_sensing}. Current research focuses on the aerodynamics of seeds that have exceptional wind dispersal mechanisms such as, for example, the \textit{Taraxacum} (dandelion)~\cite{Cummins2018Dandelion} and \textit{Alsomitra} (Javan cucumber). This case study focuses specifically on \textit{Alsomitra}-inspired drones (Figs.~\ref{fig:also},\ref{fig:overview}) as the aerodynamics underlying the flight of this diaspore is unique in the plant kingdom, enhancing the dispersal mechanism provided by the wind by an efficient gliding flight. This allows one of the heaviest seeds (314~mg)~\cite{azuma1987alsomitra} to reach a similar descent velocity than some of the lightest seeds such as the dandelion (0.6~mg)~\cite{Cummins2018Dandelion}. Because of this unique feature, several authors have considered this diaspore as a bioinspiration for microdrones~\cite{alsomitra_sensing, alsomitra_agriculture}. Such drones could function as distributed sensors in the atmosphere, for weather monitoring or detecting pollutants~\cite{electronic_microfliers, Iyer2022Wind, johnson2023origami, alsomitra_sensing, alsomitra_agriculture}. This could be particularly useful for environmental monitoring and meteorology, with research and regulations moving towards incorporating drone observations to improve weather predictions~\cite{ercdandidrone, UAS_WMO}. It has been demonstrated that such systems are capable of sustained flight with active control and internal electronics~\cite{johnson2023origami}, although more work on effective actuation and control methods is needed in the future.

Neural networks (NNs) have been widely investigated for drone control, for both quadcopters~\cite{amer2021deep, arch_comp_nn} and fixed-wing designs~\cite{richter2024RL,wada2021uav}. The control of small passive gliders is a relatively unexplored field, with the most relevant works involving larger aircraft~\cite{abouheaf2019RL, wada2021uav} or without continuous control~\cite{johnson2023origami}. For our application, we consider NN control since it has been shown to achieve accurate and robust control for systems with uncertain dynamics ~\cite{li2019control}, it is particularly applicable to controlling swarms~\cite{qamar2022swarm}, and improvements to low-order aerodynamic modelling~\cite{Li2022model} facilitates easier simulations of such drones. This approach could facilitate particularly lightweight and low-cost drones - such as with analogue network circuits printed on flexible substrates acting as the body of the gliding drone~\cite{Singaraju2022, Oshima2023}. One could alternatively consider uncontrolled flying sensors~\cite{electronic_microfliers,alsomitra_sensing,Iyer2022Wind} or traditional approaches such as state-space or model predictive control. However, one should consider that such systems will need to be verifiably safe with regard to people, other air users, and the environment~\cite{UAS_WMO}. These drones could collide with each other, veer into unsafe airspace, fall into an endangered ecosystem, or otherwise cause harm. The utility of uncontrolled flyers would be hampered by such issues, unless they can be made biodegradable. Traditional control methods may be applied, but NN methods have advantages in that they can be made data-driven and adaptive, and printed NN circuits could lead to lighter designs than digital microcontrollers.

% CHALLENGES
\subsection{Contributions}
Our first aim is the introduction of a novel case study (outlined in Sect.~\ref{sec:v_task}) in the verification of \textit{Alsomitra}-inspired drone controllers (our modelling methods are explained in Sect.~\ref{sec:alsomitra}), that differs significantly from existing benchmarks. Unlike VNN-COMP benchmarks such as ACAS Xu, our study involves regression control and continuous dynamic equations. Compared to ARCH-COMP benchmarks such as QUAD, our system involves differential equations that are far more complex in terms of the number of non-linear terms. Moreover, unlike the majority of ARCH-COMP cases, this problem does not have as natural a notion of the start, goal, safe, and unsafe states; and thus requires an out-of-the-box approach to property specification.

We propose our ideal formalisation of the problem in Sect.~\ref{sec:formal}, and distil the formalisation down to properties that can be handled with available tools (Marabou~\cite{Katz2019} implemented with Vehicle~\cite{vehicle_tutorial}, and CORA~\cite{CORA2025}) in Sects.~\ref{sec:vehicle} and ~\ref{sec:cora}. 
The choice is motivated by the fact that each can be seen as a representative of a set of tools that come from the research communities of VNN-COMP~\cite{vnn_comp} and ARCH-COMP~\cite{arch_comp}, respectively.
We present a new implementation of adversarial training for Lipschitz robustness applied to regression training for our controllers in Sect.~\ref{sec:adversarial_training}, and present the results of verifying those properties with our robust networks in Sects.~\ref{sec:v_results} and ~\ref{sec:c_results}.

Our second aim is to present the lessons learnt from investigating this case study, to help inform the development of relevant tools for similar real-life cyber-physical projects in the future (Sect.~\ref{sec:discussion}). 
%We assume that other real-life use cases of cyber-physical system verification likewise will encounter safety properties that involve highly non-linear systems with infinite or undefined planning horizons. 
The main lesson learnt is that no single existing tool ticks all the desirable boxes. Moreover, each individual tool we chose would benefit from further development in several aspects that are crucial for real-life models.
%As a consequence, they will require multiple tools to express and verify such properties, and our modifications to existing tools will be very relevant. 
Concretely:
\begin{itemize}
\item On the Vehicle side,
the verification properties that arise in the presented study are more complex than the usual VNN-COMP benchmarks in at least three ways.
\begin{itemize}
\item Firstly, the constraints on the input vector are more complex: instead of constraining individual vector elements by constants (as e.g. in $a \leq x_i \leq b$), as is the case in the majority of benchmarks including ACAS XU~\cite{katz2017reluplex}, the constraints establish relation between different vector elements, as e.g. in $x_i \leq c x_j$. This changes mathematical interpretation of the verification problem: it no longer boils down to defining a hyper-rectangle (or other constant shape) on the input space and propagating it through the network layers, but gives a more general case of linear programming that works on arbitrary input space constraints. Not every VNN-COMP~\cite{vnn_comp} verifier will be able to deal with such verification properties: Marabou is one of the most general tools in this family of tools and this case study suggests this generality may play a bigger role in the future. 
\item Secondly, for verification of Lipschitz robustness, we implemented \emph{relational properties}, i.e. properties that compare different outputs of a neural network. These properties are not natively supported by Marabou or Vehicle yet, and required some additional plumbing. On-going implementation of support for relational verification in Marabou will be useful for cases such as this. 
\item Finally, some novelty of our verification approach is derived from the fact that, unlike most benchmarks in VNN-COMP, our models are regression models, rather than verification models. Some of the methods for training and verification are specialised to classification tasks only, and we predict that this has to change with occurrence of new engineering-inspired benchmarks. 
\end{itemize}

\item On the CORA side, the system outlined in this study required several workarounds in order to compute reachability:
% \note{Can you follow the same style and bullet by bullet formulate what are the lessons learnt, in terms of comparing our bencmark to others in ARCHCOMP?}
\begin{itemize}
\item The complexity of our system of equations~\cite{Li2022model} far exceeds that of all ARCH-COMP~\cite{arch_comp} benchmarks, in the number of non-linear terms. This would cause the Jacobian and Hessian matrices to far exceed the maximum number of terms supported by MATLAB , and fail to run. The equations were simplified (Sect.~\ref{equations}) by constraining the pitch angle and using an angle-of-attack definition, solving the complexity issue, but (for any reasonably large initial set) the reachable set still tended to expand exponentially after relatively few timesteps. This was solved by dividing the initial set into smaller subsets, computing reachable sets for each, and combining the results.
\item CORA expects a NN controller that takes the system variables as inputs, with relatively few layer types supported~\cite{CORA2025}. Certain parameters occupy wider ranges than others (for example, $\theta \in [-0.93, -0.07]$, $x\in [0.48, 41.7]$) but unlike Vehicle, input normalisation (keeping all inputs between 0 and 1) is not supported. This is problematic since we intend to observe the effect of adversarial training, for which the input ranges should to be normalised, such that PGD attacks occur in $\epsilon$ ranges that are not imbalanced between input dimensions. A workaround was found by training an adversarial network on normalised data, then implementing normalisation layers to the start and end of the network.
\item Unlike similar ARCH-COMP benchmarks such as QUAD, our notion of a goal region is less obvious. We want the drone to adhere to the target trajectory in $x$ and $y$, so define a goal state as a region around that trajectory.
\end{itemize}

% Implementation of our system in CORA was particularly difficult, and we discuss the workarounds required for it to function.

% and modify normalised networks to handle unnormalised data such that robust networks can be verified in CORA.

\end{itemize}

Although this study considers only one modification of gliding drones, most of the paper's conclusions will be common between their different modifications, such as e.g. dandelion-inspired drones, and the lessons learnt can be broadly applied to other continuous control tasks. All relevant files are publicly available \href{https://github.com/ckessler2/SAIV_2025_Alsomitra}{here}.

\section{Background}
\subsection{Neural Network Control}\label{sec:nnc}
\begin{figure*}[h]
    \centering 
    \includegraphics[width=0.7\textwidth]{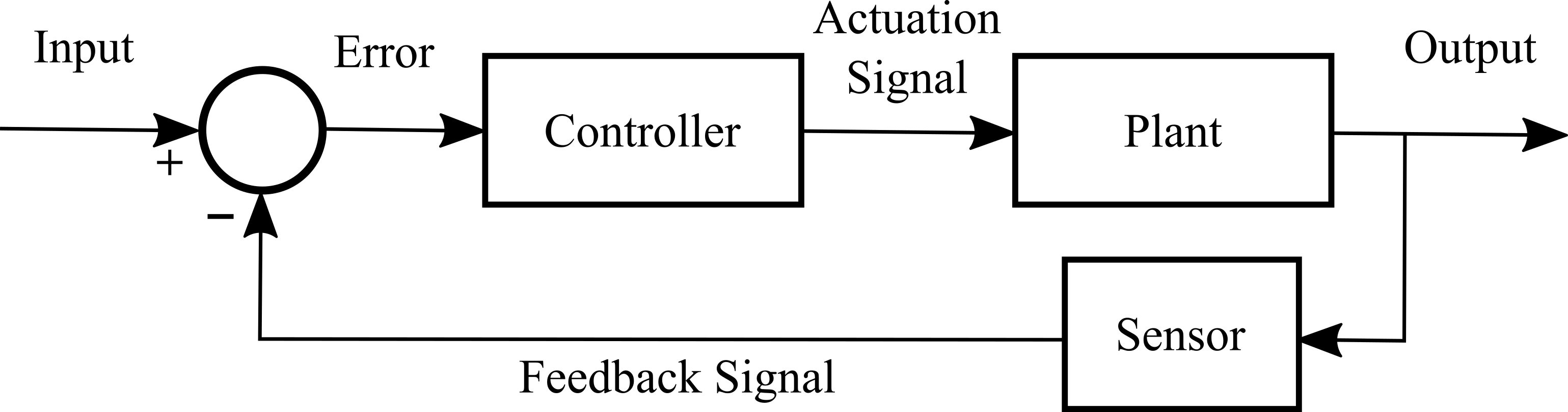}
    \caption{{{Overview of a negative feedback control system. For each control iteration, an error signal is calculated by subtracting the current system state (feedback) from the desired system state (input). A controller computes an actuation based on this error, which is applied to a simulated or real system (plant), resulting in some new output state.}}}
    \label{fig:control_overview}
\end{figure*}

For our control method, we will use the common closed-loop negative feedback method, an overview of which can be seen in Fig.~\ref{fig:control_overview}. In simple terms, the controller in a drone is given information about its current state (such as position, relative to some desired state) as input, and outputs a command to an actuator which affects how the drone flies. The controller can be considered as an equation linking the system states to an actuation force that changes the states over time according to the system dynamics, where the controller design affects how the drone behaves. If a traditional control theory approach is difficult (such as if the dynamics are highly complex) or a data-driven approach is desireable (if collecting data is easier than modelling the system, or if adaption based on new data is required), an engineer might consider implementing a NN controller.
\subsection{Verification Tools}\label{sec:tools}
The case study will rely on the following three groups of neural network verification (NNV) tools. %known within the VNN community.
The first group concerns verification of infinite time-horizon properties of controllers in isolation from verification of the overall system dynamics. The most famous benchmark in the domain is ACASXu, and the representative verifier is Marabou~\cite{Katz2019}; other tools, such as ERAN~\cite{ERAN}, Pyrat~\cite{PyRAT} or $\alpha\beta$-CROWN~\cite{ab-CROWN} could be interchangeably used for the verification tasks in which Marabou is deployed in this paper; we refer the reader to VNN-Comp~\cite{vnn_comp} for an in depth discussion of existing tools in this category.
In addition, we use Vehicle~\cite{vehicle_tutorial}, a higher-level interface on top of Marabou, and take advantage of its facility in bridging the \emph{embedding gap}~\cite{cordeiro2025neuralnetworkverificationprogramming} between the physical domains and vector representation of data. 

\begin{figure}[h]
  \begin{center}
    \includegraphics[width=0.5\textwidth]{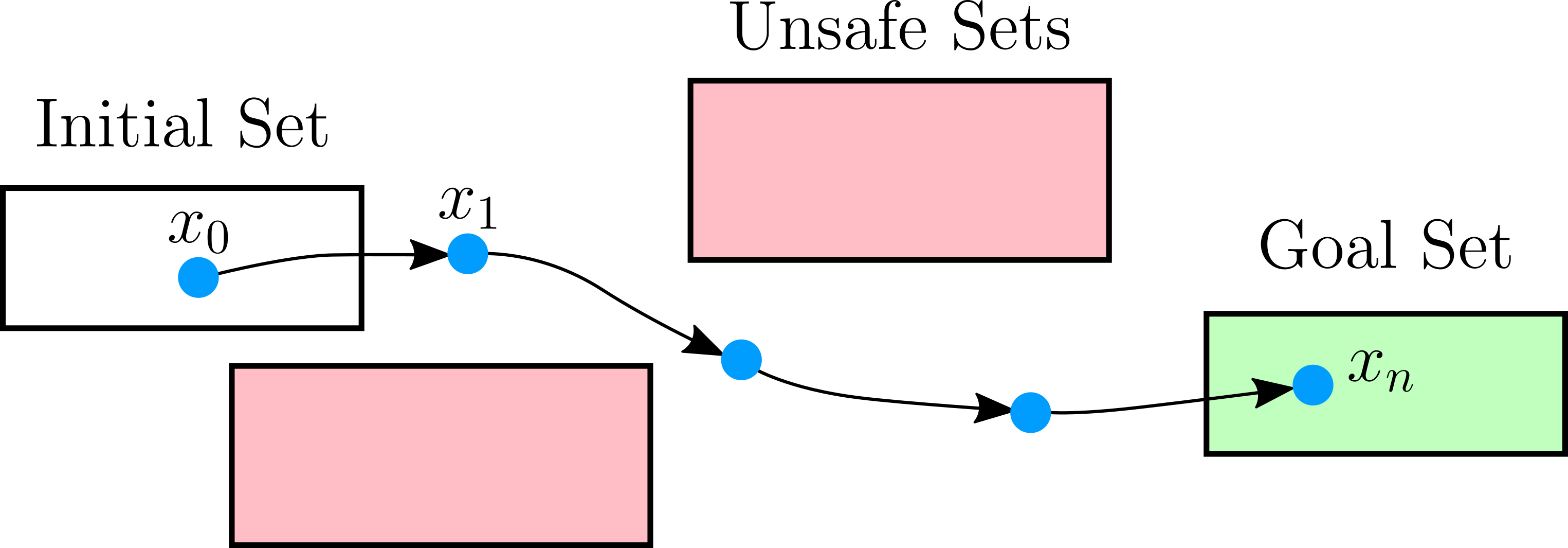}
  \end{center}
  \vspace{-0.2cm}
  \caption{{{General form of reachability specifications - dots represent the system at successive control time steps, and arrows represent the continuous trajectory of the system. Any trajectory starting in the initial set should never intersect an unsafe set, and always finish in the goal set.}}}\label{fig:reach}
  \vspace{-0.2cm}
\end{figure}
The second group of methods considers the neural controller together with the overall system dynamics to ensure that the entire system avoids unsafe states, see Fig.~\ref{fig:reach}. This class of problems is also known under the umbrella term reachability verification and representative examples include e.g. POLAR-Express~\cite{polar_express}, and CORA~\cite{CORA2025}, see~\cite{arch_comp_nn} for an exhaustive overview of the mainstream tools in this category. 
Representative benchmarks include simple dynamic problems such as the inverted pendulum, and more complex problems such as the quadcopter, space docking, and 2-wheeled obstacle avoidance. Each benchmark has a predetermined set of dynamic equations and a NN controller, with a mix of supervised learning (through behaviour-cloning) and reinforcement learning. The limitations of these benchmarks are in the complexity of the networks (large networks require reduction methods), complexity of the equations (systems are either linear, or relatively simple non-linear differential equations), and verification of complex properties (no tools can successfully verify the Spacecraft Docking benchmark as of the most recent results~\cite{arch_comp}).

Finally, an important group of methods for practical NNV cases comes from machine learning domain, under the umbrella term of \emph{property-driven training (PDT)}. These methods allow to optimise a given neural network for satisfying a desired verification property, with a view of improving the verification success~\cite{fischer2019dl2,casadio2022,flinkow2025}. Although methods in this group vary, they usually deploy a form of training with projected gradient descent (PGD)~\cite{KM18}. PGD methods involve finding the worst-case perturbed example in a region around a data point, which can then be implemented as a loss function during training:

% PGD improves the network's robustness by minimising the error associated with the \textit{worst-case} perturbed example:

\[
\min_\theta \; {E}_{(x, y) \sim \mathcal{D}} \; \left[ \max_{\delta \in \Delta} \; \mathcal{L}(f_\theta(x + \delta), y) \right]
\]

\noindent where
%\begin{itemize}
    %\item
    \(\theta\) represents the parameters of the NN;
    %\item 
    \((x, y) \sim \mathcal{D}\) are input-label pairs sampled from the data distribution \(\mathcal{D}\);
    %\item 
    \({E}\) is the expected value, averaging the loss over all samples in the data distribution \(\mathcal{D}\);
    %\item 
    \(\delta \in \Delta\) is the adversarial perturbation constrained within a feasible set \(\Delta\) (e.g., \(\|\delta\|_p \leq \epsilon\)) and
    %\item 
    \(\mathcal{L}\) is the loss function (e.g., RMSE, MAE) measuring the discrepancy between the predicted output \(f_\theta(x + \delta)\) and the true label \(y\).

The inner maximisation, which identifies the \textit{worst-case} adversarial perturbation \(\delta \in \Delta\), is performed using PGD that iteratively adjusts \(\delta\) by ascending the gradient of the loss function with respect to the input, followed by projection back onto the feasible set \(\Delta\) (e.g., ensuring \(\|\delta\|_p \leq \epsilon\)). 
 
The outer minimisation, aimed at optimising the neural network parameters \(\theta\) to minimise the adversarial loss, is achieved using gradient descent. 

\section{Modelling Methodology}\label{sec:alsomitra}
\subsection{\textit{Alsomitra macrocarpa}}
\begin{figure*}[h]
    \centering \vspace{6mm}
    \includegraphics[width=1\textwidth]{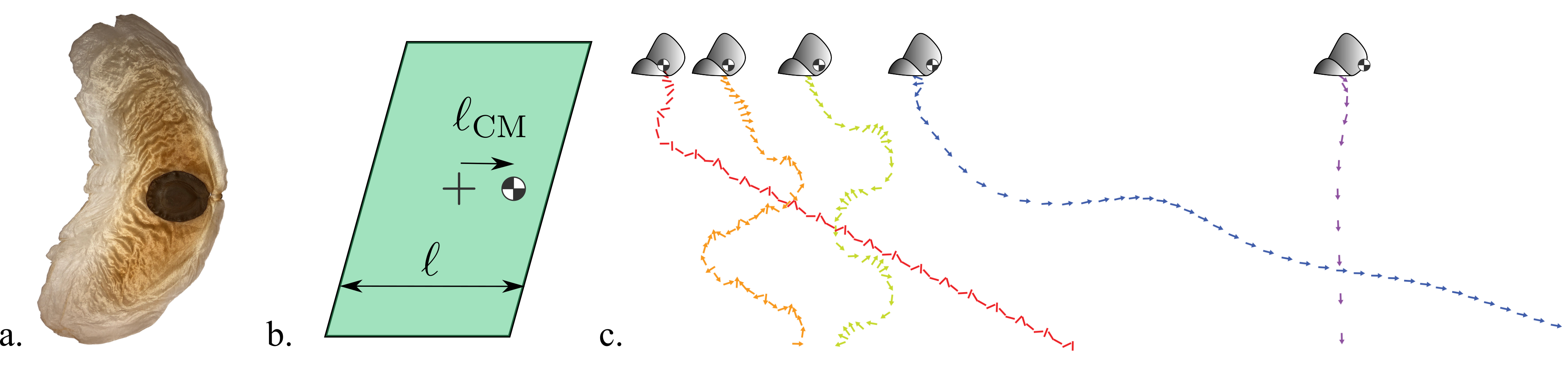}
    \caption{{{(a) An \textit{Alsomitra} seed~\cite{certini2023alsomitra}. (b) A two-dimensional approximation of an \textit{Alsomitra} seed, with centre of mass (CoM) displaced by $\ell_{\mathrm{CM}}$ (nondimensional form $e_x = \ell_{\mathrm{CM}}/\ell$). (c) Effect of various $e_x$ on gliding trajectories; according to a quasi-steady aerodynamic model (~\cite{Li2022model}, Sect.~\ref{equations}). As the CoM is displaced the trajectory behaviour is affected significantly.}}}
    \label{fig:overview}
\end{figure*}

% \textit{Alsomitra macrocarpa} seeds have a unique shape resulting in stable flight over long distances~\cite{certini2023alsomitra}, providing our inspiration for gliding drones. 
A dynamics model (Fig.~\ref{fig:overview}) was derived from~\cite{Li2022model} resulting in a system of equations for falling plates with displaced centre of mass (CoM), as defined in Sect.~\ref{equations}. Based on experimental measurements, our model accurately describes the falling trajectories of \textit{Alsomitra} seeds by inferring aerodynamic forces from the angle of attack\cite{Li2022model}. The flight characteristics are highly dependant on the CoM displacement ($e_x$, Fig.~\ref{fig:overview}), providing us with a convenient actuation method for an \textit{Alsomitra}-inspired drone. 
\input{equations}\label{equations}

\newpage\section{Verification Task} \label{sec:v_task}
Our \textit{Alsomitra} model from Sect.~\ref{equations} is used as the basis of a feedback control system with a NN controller, as described in Sect.~\ref{sec:nnc}. In our case, the plant is the aerodynamic model, and the desired input is a linear reference trajectory in $x_5$ and $x_6$ (translational $x$ and $y$):
 \begin{equation}  \label{traj_equation}
     x_6 = -x_5
 \end{equation}
 
\begin{figure}
  \begin{center}
    \includegraphics[width=0.21\textwidth]{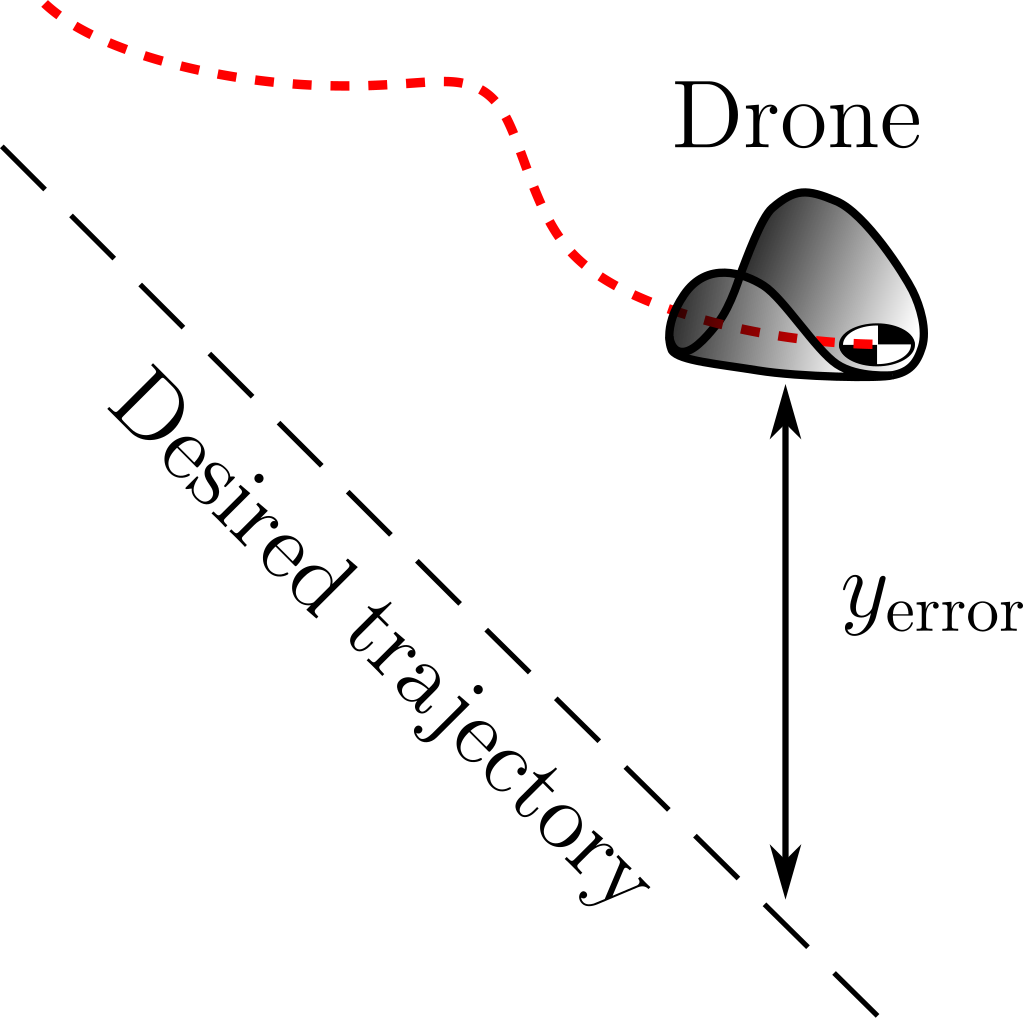}
  \end{center}
  \caption{As a control problem, we consider an \textit{Alsomitra}-inspired microdrone and attempt to follow a linear trajectory in two dimensions.}\label{fig:trajectory_diagram}
\end{figure}

 The feedback signal consists of the six system states, and the CoM displacement is actuated by a controller aiming to follow the target trajectory (Fig.~\ref{fig:trajectory_diagram}). 

\begin{figure}[h] 
 \begin{center}
 \includegraphics[width=0.65\textwidth]{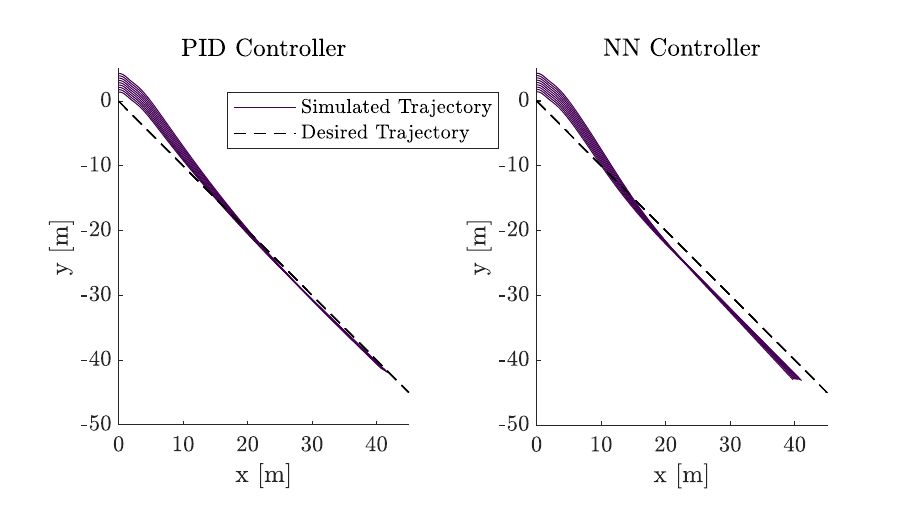}
    \caption{{{PID and basic NN controller performance on an \textit{Alsomitra}-inspired drone. The naive network is trained on regression data obtained from simulations with the PID controller, and the resulting performance is similar but not perfect.}}}
    \label{fig:trajectory_plots}
 \end{center}
\end{figure}

 As per the ARCH-COMP airplane and pendulum benchmarks~\cite{arch_comp}, the neural network controller is trained using behaviour cloning. All simulations ran for a total of 20~s, with a model timestep of 0.01~s and a control timestep of 0.5~s. A PID controller actuates $y_1$ ($e_x$) based on an error in $x_6$, and the gains are tuned manually until the control system performs well for a range of starting $x_6$ positions. 
For each controller actuation (24 per simulation, for nine simulations), the system states, $x_6$ error, and PID actuation are recorded for use as training data. This data is imported to Python for standard regression learning, and networks are exported in .onnx format for evaluation (Fig. ~\ref{fig:trajectory_plots}) and verification. All networks have 6 inputs, 3 hidden layers with 6, 4, and 1 nodes respectively with ReLU activation functions, and 1 output.

 \subsection{Formalisation}\label{sec:formal}
 The core of this case study lies in examining the challenges in adopting the existing NNV methods in this new domain. Our ideal formalisation of the problem would be as follows. We consider a hybrid program where the six system states $x_{1},...,x_6$ change over continuous time $t$ according the dynamics model shown in Sect.~\ref{equations}, and a NN controller acts to change the system state discretely every 0.5~s. For any starting state $x_{1},...,x_6(0)$, after some time $t^*$ the trajectory of the drone will always be within some small distance $y^*$ of the target trajectory (ideally, $x_6=-x_5$). This boils down to the following ideal verification property:

\begin{equation} \label{eq:specific_trajectory_robustness}
	\forall t \geq t^*, \forall x_{1},...,x_6(0) \in \mathbb{R} : |x_6(t) + x_5(t)| \leq y^*
\end{equation}

There are several features that distinguishes this system from standard NNV benchmarks, and we aim to explain the technical implications of these challenges for existing verification technologies, and propose ways in which these challenges can be overcome:

\begin{enumerate}
\item The system dynamics are continuous, therefore unlike standard control verification benchmarks (such as ACAS Xu~\cite{vnn_comp}), control is modelled as a regression task as oppose to classification.
\item Unlike the ARCH-COMP benchmarks~\cite{arch_comp_nn} that have a pre-defined notion of safe and unsafe state, these gliding drones do not have a notion of safety in the sense of a pre-defined coordinate region. A safe state is instead defined in a relational way, as adhesion to certain safe trajectory.
\item Unlike many ARCH-COMP benchmarks, our verification task requires modelling with an infinite time horizon. Each drone could stay airborne for an arbitrary duration of time, depending on the surrounding airflow.
\item As defined by our model, the dynamics of gliding drones are more complex than what is currently handled by the ARCH-COMP benchmarks and tools, and in particular it is more complex than the dynamics handled by tools that can verify infinite-time horizon systems, such as KeyMaeraX~\cite{Platzer18}.
\end{enumerate}

The available verification tools (Sect.~\ref{sec:tools}) do not allow us to formalise this idealised goal directly, since CORA does not support infinite time, and Marabou does not support differential equations. As a result, we simplified this general task as two simpler tasks (in the first case sacrificing the analysis of the overall system dynamics, and in the second case the infinite-time horizon and relational notion of the target state):
\begin{enumerate}
\item The NN will never command the drone to deviate significantly from the target trajectory. This task was implemented in Marabou, using the Vehicle specification language since it facilitates complex property definition.\label{task1}
\item Given an interval of initial positions and a finite time horizon, the NN-controlled drone will always reach a goal region, defined as a region around the target trajectory within this finite time frame. This task was verified in CORA.\label{task2}
\end{enumerate}
We note that task 1 resembles in some way a robustness property~\cite{casadio2022}, except for now we deal with a regression NN and robustness relative to a line rather than a given data point.

\section{Robustness Training for Regression}\label{sec:adversarial_training}
Since robustness is critically important for drone safety, it seems reasonable to attempt a form of adversarial training based on PGD methods for our controller. The guiding hypothesis was that a general improvement in NN robustness should lead to improved verification performance. Since our neural network is a regression model, the classification-based training methods surveyed in Sect.~\ref{sec:tools} could not easily be used without modification. We therefore had to modify the PGD algorithm to use an RMSE loss function instead of cross-entropy, and modify other aspects of the adversarial training algorithm that relied on the presence of discrete classes. 
%that  is employed as the loss function. 

We focused on two notion of robustness, \emph{standard} and \emph{Lipschitz} robustness~\cite{casadio2022}. Given $x^* \in \mathcal{D}$ and constants $\epsilon, \delta, L \in R$, 
\begin{equation} \label{eq:robustness}
	\forall x \in {R}^n : \|x - x^*\| \leq \epsilon \implies \|f(x) - f(x^*)\| \leq \delta
\end{equation}
\begin{equation} \label{eq:lipschitz_robustness}
	\forall x \in {R}^n : \|x -x^*\| \leq \epsilon \implies \left\| f \left ( x \right ) - f\left ( x^* \right )  \right\|  \leq  L \left\| x - x^* \right\|
\end{equation}

Since the latter has been proven to be strictly stronger than the former in~\cite{casadio2022}, we implemented a form of PGD training with a Lipschitz loss function. During each training epoch, the algorithm finds the worst-case adversarial example ($x^*$, $f(x^*)$) in an $\epsilon$-ball around each training point ($x$, $f(x)$).To optimise the regression model for Lipschitz robustness, we dynamically compute the highest value of $L$ from the training and adversarial points (according to Eq.~\ref{eq:lipschitz_robustness}), which is summed to the training RMSE loss, penalising the network for large gradients about each data point. This is expected to result in a network with a smoother and therefore more robust output, at the expense of some accuracy.

\section{Property-Driven Training (PDT)}\label{sec:pdt}
% For a system such as ours, a compelling option is to incorporate verification into training in order to improve performance and robustness. A training method using differentiable logics applied to this case study has been investigated in~\cite{flinkow2025}, and certain PDT NNs are shown in Table~\ref{tab:prop123}. 

A different PDT method using differentiable logics was independently developed and applied to this case study in a separate submission~\cite{flinkowGeneralisedFramework2025}. As that submission does not include verification experiments, we will evaluate several models optimised for Properties 1, 2, 4, and 5 in ~\cite{flinkowGeneralisedFramework2025} for the sake of comparison. The models are listed in Table~\ref{tab:prop123}.
% The models trained as described in Sect.~\ref{sec:adversarial_training} are named \textit{Adversarial}, and models trained using the optimisation algorithm of ~\cite{flinkowGeneralisedFramework2025} are named \textit{DL2} and \textit{Godel Logic} respectively. 
In the table, \textit{adversarial} results represent just by one model trained according to Sect.~\ref{sec:adversarial_training}; but \textit{DL2} and \textit{Gödel Logic} models are optimised specifically for Properties 1, 2, 4, and 5 respectively.
% \note{Suggestion:  A different property-driven training method using differentiable logics was *independently* debveloped and applied to this case study in a separate submission~\cite{flinkowGeneralisedFramework2025}. As that submission did not include verification experiments, for the sake of comparative evaluation, we will evaluate several models optimised for Properties 1-5 (outlined in Sect.~\cite{sec:vehicle}) in ~\cite{flinkowGeneralisedFramework2025}.  The models are listed  in Table~\ref{tab:prop123}. The models trained as described in Section~\ref{} are named ``Adversarial", and models trained using the optimisation algorithm of ~\cite{flinkowGeneralisedFramework2025} are named ``DL2" and "Godel Logic", respectively. }
A value of $y^*=2$ was chosen for properties 1, 2, and 4 in training, in order to keep the properties relatively strict whilst avoiding counterexamples in the dataset.

\section{Vehicle Implementation}\label{sec:vehicle}
Task~\ref{task1} was broken down into five simpler specifications to be implemented in Vehicle (a detailed introduction to which can be found in~\cite{vehicle_tutorial}), to ensure the NNs control the drone as desired in various ways. Global properties relating to the controller's output relative to the target trajectory are introduced in Sect.~\ref{sec:global_properties}, and a local robustness property is introduced in Sect.~\ref{sec:local_property}. Global properties are verified for all inputs bounded by the training data (representing the entire parameter space over which our controller is trained), and the local property is evaluated about $\epsilon$-balls from the traiing data.
% \knote{I suggest you have a detailed discussion about the Vehicle code: firstly, would be good to explain general housekeeping -- variable declarations. normalisation, max and min values etc; then -- any intricaies of encoding or evaluating the main properties. Do you evaluate any of them locally using Vehicle facilities for idx files? These are all worth mentioning -- these details will be relatively new for SAIV community that is used to VNN-LIB formats.}
\subsection{Global Property Specifications}\label{sec:global_properties}
Our first goal is to ensure the controller never causes the drone to deviate from the target trajectory. To establish a performance criteria, properties 1-4 include $y^*$ (a threshold distance from the target trajectory, Eqs.~\ref{traj_equation},~\ref{eq:specific_trajectory_robustness}), such that a critical $y^*$ can be found per network per property where verification succeeds. For example, for properties 1 and 2, a lower critical $y^*$ would indicate a controller that better adheres to the target trajectory:

\begin{enumerate}
    \item \textit{If the drone is above the line by some threshold $y^*$, the NN output will always make the drone pitch down} (Listing~\ref{listing:1})
    \begin{equation}\label{eq:prop1}
        x_6 \ge -x_5+y^* \Rightarrow f(x) \ge 0.187
    \end{equation}
    
    \item \textit{If the drone is below the line by some threshold $y^*$, the NN output will always make the drone pitch up}
    \begin{equation}\label{eq:prop2}
        x_6 \le -x_5-y^* \Rightarrow f(x) \le 0.187
    \end{equation}
\end{enumerate}

Our third property is reversed, where a larger $y^*$ would indicate better adherence to a larger region around the target trajectory:

\begin{enumerate}[start=3]
    \item \textit{If the drone is close to the line by some threshold $y^*$, and at an intermediate pitch angle, the NN output will always be intermediate} (Listing~\ref{listing:3})
    \begin{equation}\label{eq:prop3}
        -x_5 - y^* \leq x_6 \leq -x_5 + y^* \land -0.786 \leq x_4 \leq -0.747 \Rightarrow 0.184 \leq f(x) \leq 0.19
    \end{equation}
\end{enumerate}

Our fourth property is more complex, and represents a desireable behaviour not present in the data:

\begin{enumerate}[start=4]

    \item \textit{If the drone is above and close to the line, pitching down quickly and moving fast, the NN output will always make the drone pitch up}
    \begin{equation}\label{eq:prop4}
        -x_5 \le x_6 \le -x_5+y^* \hspace{2mm} \land \hspace{2mm} x_3 \le -0.12 \hspace{2mm} \land \hspace{2mm} x_2 \le -0.3  \Rightarrow f(x) \le 0.187
    \end{equation}
\end{enumerate}

In our Vehicle code, $alsomitra$ represent the NN, $validInput$ represents the input space bounded by the training data, and the parameter $ystar$ is defined during runtime.

% \knote{please have a look at listing 1 -- I added the declaration of Haskell as a language and it has better highlighting -- generally minted supports a bunch of programming languages. Also: consider using Figure instead of listing, and place it on top of pages}
\begin{listing}[t] 
\caption{Property 1 implemented in Vehicle.}
\begin{minted}[fontsize=\scriptsize]{haskell}
droneFarAboveLine : UnnormalisedInputVector -> Bool
droneFarAboveLine x =
	x ! d_y >= - x ! d_x + ystar

@property
property1 : Bool
property1 = forall x . validInput x and droneFarAboveLine x =>
  alsomitra x ! e_x >= 0.187
\end{minted}
\label{listing:1}
\end{listing}

% \begin{listing}[t] 
% \caption{Property 2 implemented in Vehicle}
% \begin{minted}[fontsize=\scriptsize]{haskell}
% droneFarBelowLine : UnnormalisedInputVector -> Bool
% droneFarBelowLine x =
% 	x ! d_y <= - x ! d_x - ystar

% @property
% property2 : Bool
% property2 = forall x . validInput x and droneFarBelowLine x =>
%   alsomitra x ! e_x <= 0.187
% \end{minted}
% \label{listing:2}
% \end{listing}

\begin{listing}[!t] 
\caption{Property 3 implemented in Vehicle.}
\begin{minted}[fontsize=\scriptsize]{haskell}
intermediatePitch : UnnormalisedInputVector -> Bool
intermediatePitch x =
	-0.786 <= x ! d_theta <= -0.747

closeToLine : UnnormalisedInputVector -> Bool
closeToLine x =
	x ! d_y >= -x ! d_x - ystar and
	x ! d_y <= - x ! d_x + ystar
    
@property
property3 : Bool
property3 = forall x . validInput x and intermediatePitch x 
and closeToLine x => 0.184 <= alsomitra x ! e_x <= 0.19
\end{minted}
\label{listing:3}
\end{listing}

\subsection{Local Robustness Specification}\label{sec:local_property}
Our fifth property is an evaluation of robustness around $\epsilon$-balls with respect to the training dataset, as defined in Sect.~\ref{sec:adversarial_training}. A detailed introduction to $\epsilon$-ball robustness for image classification implemented in Vehicle can be found in~\cite{vehicle_tutorial}. Similarly to properties 1-4, we are interested in finding at what threshold $L$ value ($L^*$) does each network pass verification. Similarly to properties 1-4, we expect the verification results for this property to depend on the strictness of $L$, which we consider as the parameter $L^*$.
However, due to Marabou limitations, this was evaluated with respect to the training dataset, where for each network Property 5 was evaluated for each training point, given fixed $L*$ and $\epsilon$ values. Additionally, the distance between points was computed with $L^\infty$ and the input distance could not be included in the formula, leading use to use a different robustness definition:
\begin{enumerate}[start=5]
    \item \textit{For any given input point $x$, the network output $f(x^*)$ of any perturbed point $x^*$ within an $\epsilon$-ball around $x$, will have a distance less than or equal to $L^*/\epsilon$ to $f(x)$} (Listing~\ref{listing:4})
    \begin{equation}\label{eq:prop5}
        \forall x \in {R}^n : \|x - x^*\| \leq \epsilon \implies \|f(x) - f(x^*)\| \leq L^*/\epsilon
    \end{equation}
\end{enumerate}
This definition is less strict than our definition of Lipschitz robustness (Eq.~\ref{eq:lipschitz_robustness}), since it is effectively equivalent to standard robustness (Eq.~\ref{eq:robustness}) where $L^*/\epsilon=\delta$. This means that any counterexample to Property 5 will also violate Lipschitz robustness where $L = L^*$, but not the other way around. Since we train for the stronger definition (Sect.~\ref{sec:adversarial_training}), we expect to see improved robustness with regard to this weaker definition.

In our Vehicle code, parameters $epsilon$ and $Lipschitz$ are defined during runtime, and $n$ is inferred from the training data (provided in idx format~\cite{vehicle_tutorial}).

% \knote{Again here: Vehicle code needs discussion, it is not something that each SAIV reviewer finds familiar. }
% \knote{see also my questions on slack for other questions that came to my head}
% \knote{when evaluating Property 5 and its "foreach", it is worth pointing out that Vehicle enables this easy syntax vis the declaration "@parameter(infer=True)
% n : Nat" and idx}
% \knote{captions to figures or listings better be in footnotesize, as you set the minted code to footnotesize... and they contrast}

\begin{listing}[h] 
\caption{Property 5 implemented in Vehicle. In this case, the states are defined in normalised terms to avoid scaling issues. Instead of calling the network twice to evaluate $f(x)$ and $f(x^*)$, the NN is doubled in onnx format so that two sets of inputs and outputs can be evaluated at once.}
% boundedByEpsilon : InputVector -> Bool
% boundedByEpsilon x = -epsilon <= x ! dv_x - x ! dv_x2 <= epsilon and
% 	-epsilon <= x ! dv_y - x ! dv_y2 <= epsilon and
% 	-epsilon <= x ! d_omega - x ! d_omega2 <= epsilon and
% 	-epsilon <= x ! d_theta - x ! d_theta2 <= epsilon and
% 	-epsilon <= x ! d_x - x ! d_x2 <= epsilon and
% 	-epsilon <= x ! d_y - x ! d_y2 <= epsilon
\begin{minted}[fontsize=\scriptsize]{haskell}
myList : List Rat
myList = [0, 1, 2, 3, 4, 5]

boundedByEpsilon : InputVector -> Bool
boundedByEpsilon x = forall i in myList . -epsilon <= x ! i - x ! i + 6 <= epsilon

validPerturbation : InputVector -> Bool
validPerturbation x = forall i in myList . x ! i == 0.0

standardRobustness : InputVector -> OutputVector -> Bool
standardRobustness input output = forall pertubation .
  	let perturbedInput = input - pertubation in validPerturbation pertubation and 
	validInput perturbedInput and boundedByEpsilon perturbedInput =>
	(output ! e_x - alsomitra perturbedInput ! e_x2) <= Lipschitz / epsilon and 
	alsomitra perturbedInput ! e_x2 - output ! e_x <= Lipschitz / epsilon 

@property
property4 : Vector Bool n
property4 = foreach i . standardRobustness (trainingInputs ! i) (trainingOutputs ! i)
\end{minted}
\label{listing:4}
\end{listing} 

\newpage\subsection{Verification Results}\label{sec:v_results}
% \vspace{-5mm}
\begin{table}[h]
\begin{center}
\caption{Critical $y^*$ values for properties 1-4 (Sect.~\ref{sec:global_properties}), for naive and adversarially trained NNs (Sect.~\ref{sec:adversarial_training}), and for PDT NNs (Sect.~\ref{sec:pdt}). For properties 1, 2, and 4, a lower $y^*$ indicates a controller that better adheres to the target trajectory, and the inverse for Property 3.}
% \begin{tabular}{@{}ccccc@{}}
% \toprule
% \multicolumn{1}{l}{\hspace{1mm}\textbf{Property}\hspace{1mm}} & \multicolumn{1}{l}{\hspace{1mm}\textbf{Naive NN}\hspace{1mm}} & \multicolumn{1}{l}{\hspace{1mm}\textbf{Adversarial NN}\hspace{1mm}} \\ \midrule
% 1 (Eq.\ref{eq:prop1}) & {\color[HTML]{000000} 42} & {\color[HTML]{000000} 30} \\
% 2 (Eq.~\ref{eq:prop2}) & {\color[HTML]{000000} 42} & {\color[HTML]{000000} 42} \\
% 3 (Eq.~\ref{eq:prop3}) & {\color[HTML]{000000} Failed} & {\color[HTML]{000000} Failed} \\ \bottomrule\label{tab:prop123}
% \end{tabular}
\begin{tabular}{@{}ccccc@{}}
\toprule
Property\hspace{1mm} & \hspace{1mm}Naive\hspace{1mm}  & \hspace{1mm}Adversarial\hspace{1mm} & \hspace{1mm}DL2\hspace{1mm} & \hspace{1mm}Gödel Logic \\ \midrule
1 (\ref{eq:prop1})        & 46     & 30          & 30  & 27          \\
2 (\ref{eq:prop2})        & 42     & 42          & 42   & 42           \\
3 (\ref{eq:prop2})        & Failed     & Failed          &    &            \\
4 (\ref{eq:prop3})        & 0 & 0      & 0   & 0           \\ \bottomrule\label{tab:prop123}
\end{tabular}
\end{center}
\end{table}

\begin{table}[!h]
\begin{center}
\caption{Verification success rates ($\%$) of Property 5 (Sect.~\ref{sec:local_property}) for naive and adversarial NNs (Sect.~\ref{sec:adversarial_training}), per $L^*$ values and $\epsilon$, with respect to the training dataset. A higher success rate means that the NN is robust with respect to more of the training data points. As $\epsilon$ increases we are increasing the radius for perturbation around each training point, and a decreasing $L^*$ results in a stricter maximum gradient threshold. Empty cells represent properties that timed out before verifying 100 data points.
% Red cells represent properties that timed out before verifying  a significant number of data points; yellow entries completed verification for over 100 data points but not the entire dataset.
}

\begin{tabular}{@{}cccccc@{}}
\multicolumn{2}{c}{} & \multicolumn{4}{c}{$L^*$} \\ \cmidrule(l){3-6} 
\multicolumn{2}{c}{\multirow{-2}{*}{Naive}} & $\hspace{1mm}0.00001\hspace{1mm}$ & $\hspace{1mm}0.0001\hspace{1mm}$ & $\hspace{1mm}0.001\hspace{1mm}$ & $\hspace{1mm}0.01\hspace{1mm}$ \\ \cmidrule(l){3-6} 
\multicolumn{1}{c|}{} & \multicolumn{1}{c|}{0.00001} & {\color[HTML]{000000} $100$} & {\color[HTML]{000000} $100$} & {\color[HTML]{000000} $100$} & {\color[HTML]{000000} $100$} \\
\multicolumn{1}{c|}{} & \multicolumn{1}{c|}{0.0001} & {\color[HTML]{000000} $96.1$} & {\color[HTML]{000000} $100$} & {\color[HTML]{000000} $100$} & {\color[HTML]{000000} $100$} \\
\multicolumn{1}{c|}{} & \multicolumn{1}{c|}{0.001} & {\color[HTML]{000000} $17.0$} & {\color[HTML]{000000} $98.5$} & {\color[HTML]{000000} $100$} & {\color[HTML]{000000} $100$} \\
\multicolumn{1}{c|}{\multirow{-4}{*}{$\epsilon\hspace{1mm}$}} & \multicolumn{1}{c|}{0.01} & {- } & {- } & {- } & {\color[HTML]{000000} $100$}\label{tab:prop4base}
\end{tabular}

\vspace{5mm}

\begin{tabular}{@{}cccccc@{}}
\multicolumn{2}{c}{} & \multicolumn{4}{c}{$L^*$} \\ \cmidrule(l){3-6} 
\multicolumn{2}{c}{\multirow{-2}{*}{Adversarial}} & $\hspace{1mm}0.00001\hspace{1mm}$ & $\hspace{1mm}0.0001\hspace{1mm}$ & $\hspace{1mm}0.001\hspace{1mm}$ & $\hspace{1mm}0.01\hspace{1mm}$ \\ \cmidrule(l){3-6} 
\multicolumn{1}{c|}{} & \multicolumn{1}{c|}{0.00001} & {\color[HTML]{000000} $100$} & {\color[HTML]{000000} $100$} & {\color[HTML]{000000} $100$} & {\color[HTML]{000000} $100$} \\
\multicolumn{1}{c|}{} & \multicolumn{1}{c|}{0.0001} & {\color[HTML]{000000} $99.7$} & {\color[HTML]{000000} $100$} & {\color[HTML]{000000} $100$} & {\color[HTML]{000000} $100$} \\
\multicolumn{1}{c|}{} & \multicolumn{1}{c|}{0.001} & {\color[HTML]{000000} $23.9$} & {\color[HTML]{000000} $99.7$} & {\color[HTML]{000000} $100$} & {\color[HTML]{000000} $100$} \\
\multicolumn{1}{c|}{\multirow{-4}{*}{$\epsilon\hspace{1mm}$}} & \multicolumn{1}{c|}{0.01} & {\color[HTML]{000000} $0$} & {\color[HTML]{000000} $13.6$} & {\color[HTML]{000000} $92.5$} & {\color[HTML]{000000} $100$}
\end{tabular}

\end{center}
\end{table}

% \knote{In Table 1, I do not understand what values 0 stand for.}

% \knote{This section should end with a good para that interprets these results from the engineering point of view. We discussed possible ways to interpret these. The conclusions can be both positive and negative. Eg, y* being so large may render verification results useless for real applications. But the table 2 may have some nuanced interpretation -- which of its combinations look plausible from the point of view of verifying the real drones? what would be the desirable result?}

These results provide interesting insights from an engineering perspective. From Table~\ref{tab:prop123} there is a clear improvement in performance for Property 1 when implementing adversarial training and PDT, suggesting that our approaches have been successful. However, Properties 1 and 2 only succeed with very large values of $y^*$ - our controllers only adhere to a region around the target trajectory so wide as to be useless in real applications. Properties 3 and 4 failed and succeeded, for all $y^*$ values, for all the tested networks, suggesting that they are particularly difficult and easy to verify respectively. Table~\ref{tab:prop4base} shows a marginal improvement in robustness performance for our adversarial network, suggesting that our approach has been moderately successful.

\section{CORA Implementation}\label{sec:cora}
\subsection{Reachability Specification}
Since our case study is based on the QUAD benchmark from the ARCH competition~\cite{arch_comp_nn}, our reachability specification is defined similarly. The initial set is:
\begin{equation}\label{initialset}
        x_1 =1, x_2 = 0, x_3 = 0, x_4 = 0, x_5 = 0, x_6 \in [1.43,4.29]
\end{equation}
The reachability goal is for the drone to always be within a distance $y^*=2$ of the target trajectory after 20~s. To compute reachability, CORA uses set representations (such as zonotopes~\cite{CORA2025}) and set operations to over-approximate the continuous time reachable set in discrete time steps. Additional considerations include the initial set representation, time step size, controller type, and reachability algorithm - in our case we use a zonotope, 0.01~s, a NN controller, and conservative linearization.

\subsection{Reachability Results}\label{sec:c_results}
\begin{figure}[h]
  \begin{center}
    \includegraphics[width=1\textwidth]{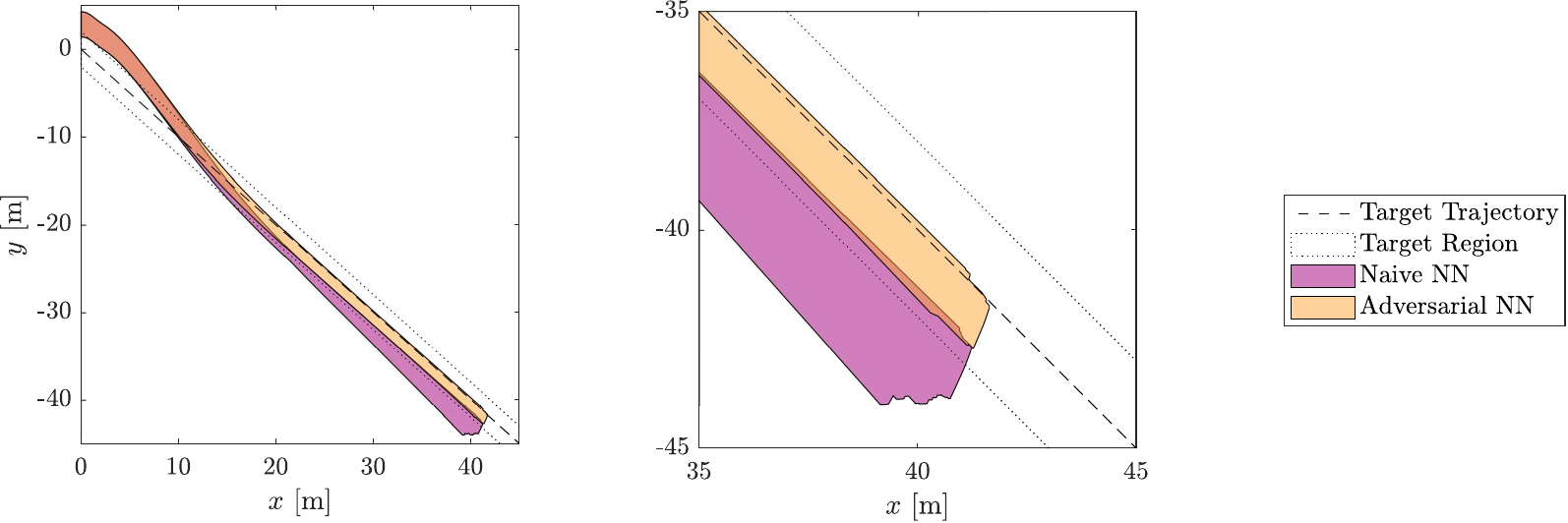}
  \end{center}
  \vspace{-0.5cm}
  \caption{Reachable regions in $x_5$ and $x_6$ for naive and adversarial NNs implemented as controllers, from the initial set defined in Eq.~\ref{initialset}. The naive NN fails to reach a region bounded by $y^*=2$ after 20~s, and the adversarial NN succeeds.}\label{fig:reach_plot}
  \vspace{-0.1cm}
\end{figure}

This result shows significant improvement from the adversarial network compared to the naive (these networks are identical to those from Tables ~\ref{tab:prop123} and ~\ref{tab:prop4base}), with the adversarial network adhering much closer to the target trajectory. This would suggest that our adversarial training has been successful, but we note that whilst these networks are identical in terms of structure, training data, and training epochs, the discrepancy in this plot could be partly due to differences in regression performance.

\section{Discussion}\label{sec:discussion}
\subsection{Limitations and Lessons Learned}\label{sec:limitations}
A use case such as ours requires multiple tools to verify, each with benefits and drawbacks. Although standard methods apply, modifications are needed for proper implementation, such as with robustness training (Sect.~\ref{sec:adversarial_training}). Vehicle does not have integration with Python, and our dynamics model and CORA are in Matlab, requiring the use of three programming languages for our case study. Marabou does not support multiple network calls, so a workaround involving doubling the NN (in .onnx format) was required to evaluate robustness in Vehicle (\ref{listing:4}). Additionally, for our local robustness property, certain Marabou queries timed out after very few data points (Table~\ref{tab:prop4base}). Correct handling of normalisation was found to be very important, since the implemented robustness training and verification methods require normalisation but CORA does not support normalisation.
To evaluate reachability in CORA with normalised networks required another workaround, incorporating the normalisation arithmetic as extra layers in the network. Furthermore, CORA could not evaluate reachability for the full system of equations described in Sect.~\ref{equations}, due to their complexity and the size of the starting set. 

Implementing CORA for our application was difficult due to set explosions, where the complexity of the reachability computation would cause an exponential expansion in the set, causing a crash. To reduce the complexity of the equations, the angle of attack definition was simplified (~\ref{eq:alpha}). To solve the initial set size issue, the initial set was divided in $x_6$, and the resulting subsets combined to a final reachable set. The reachability timesteps still needed to be small (0.01~s) to avoid set explosions; resulting in long computation times for full reachable sets (over 8 hours in some cases). All reachable sets only involve a starting interval in $x_6$ (starting values are constants in every other dimension), since increasing the starting interval size in multiple dimensions would cause set explosions. These limitations in CORA were found to stem from the complexity of the partial derivative matrices (Jacobian and Hessian), especially due to a large number of nonlinear terms. CORA could be improved significantly by faster approximations of these derivatives, such as with finite differences instead of symbolic methods, but performing such calculations over sets was found to be non-trivial. ReLU activation functions were also found to reduce computation times compared to sigmoid. 

Another consideration for future work is the trade-off between robustness and regression performance for such a controller. For our verification implementations we assume that each NN is capable of controlling the drone relatively well, but that may not always have been the case. A better comparison could be made using NNs with equivalent regression performance on a test set of data, using a coefficient such as $R^2$. Additionally, the effect of PDT~\cite{flinkowGeneralisedFramework2025} on our system was not fully investigated, partly due to differences in network structure and unresolved issues with the Marabou solver. 

% \CK{Compare to taxinet / archcomp}
% \knote{Vehicle/marabou also gave some trouble, eg their inability to handle multiple applications of a neural net. Vehicle promises but still misses integration with Python. Feel free to mention all of these.}

\subsection{Conclusion}\label{sec:conclusion}

%\section{Verification Challenges}\label{sec:challenges}
In this paper, we introduce a novel case study in the verification of gliding drones (Sect.~\ref{sec:v_task}). We developed a two-dimensional model for \textit{Alsomitra}-inspired drones in Sect.~\ref{sec:alsomitra}, presented an ideal verification formalisation in Sect.~\ref{sec:formal}, reduced the formalisation into properties manageable by current tools (Vehicle and CORA, Sects.~\ref{sec:vehicle} and ~\ref{sec:cora}), and presented verification results in Sects.~\ref{sec:v_results} and ~\ref{sec:c_results}. We noted challenges that this class of problems presents, among which are its more complex dynamics, under-defined notion of safe and unsafe states, preference for infinite-time horizon guarantees, and its different learning-as-regression regimes. We have shown that in principle, a combination of existing verification tools (Sect.~\ref{sec:tools}) and novel training methods (Sects.~\ref{sec:adversarial_training} and ~\ref{sec:pdt}) could be effectively adopted in order to enable future inclusion of this class of benchmarks into NNV portfolios. We note, however, that verification tasks like this motivate strongly development of tools that cross the boundaries of ARCH-COMP and VNN-COMP on the one hand, and incorporate these tools more smoothly with machine learning toolboxes on the other hand; cohering perhaps with the general agenda of building more complex programming language interfaces for such more complex verification tasks~\cite{cordeiro2025neuralnetworkverificationprogramming}. Technical problems such as insufficient support for normalisation (as reported here) can make a difference between verification success and failure, yet they are often over-looked in papers and tools that are dedicated to implementing NNV algorithms. Whilst we can define and verify for the behaviours that we want to a certain extent, the state of tools makes it difficult to say exactly how well NNs will fulfil their role - even for our relatively simple (two-dimensional, non-turbulent) drone system. For example, Table~\ref{tab:prop123} suggests that our NNs are only guaranteed to adhere to a very large region around the line, and Fig.~\ref{fig:reach_plot} shows reachable regions but for which the initial set is relatively small (zero-width in 5 dimensions). All of these issues will likely be found on any comparable regression task with complex dynamic equations. If these limitations can be overcome, it will help enable engineers to develop safe and robust control and modelling methods for technologies that improve people's lives and reduce our impact on the environment.

\bibliographystyle{splncs04}
\bibliography{References}

\section{Appendix}
\input{appendix}

\end{document}

%% file: equations.tex
\subsection{Equations}

The following equations describe falling plates with a displaced centre of mass~\cite{Li2022model}, with six system variables ($x_{1 ... 6}$, Equations~\ref{eq:x1} ... \ref{eq:x6}), involving mechanical ($\ell$, $m$, $g$, $\rho_f$, $I$) and aerodynamic ($C_{CP}^{0}$, $C_{CP}^{1}$, $C_{CP}^{2}$, $C_{L}^{1}$, $C_{L}^{2}$, $C_{D}^{0}$, $C_{D}^{1}$, $C_{D}^{\pi/2}$, $C_{R}$, $\alpha_0$, $\delta$) constants chosen to match that of \emph{Alsomitra} seeds~\cite{certini2023alsomitra}. Several intermediate terms are included for simplicity (Equations \ref{eq:alpha} ... \ref{eq:arr})., and a more detailed overview can be seen in Appendix~\ref{app:equations}

% When integrated over time, the force coefficients are derived from the angle of attack (5-7), the forces are calculated (8-12), followed by the equations of motion (13-18). For simplicity the plate is assumed to be infinitesimally thin ($h\approx0$), and to always have an angle of attack within the region $\alpha \in [-\pi /2, 0]$.

\begin{equation}
        \tan\alpha = (x_2 - x_3y_1\ell)/x_1\approx x_2/x_1
        \label{eq:alpha}
\end{equation}

\begin{equation}
        f = (1-\tanh((\alpha-\alpha_0)/\delta ))/2
        \label{eq:f}
\end{equation}

\begin{equation}\label{eq:CL}
        -C_{\textnormal{L}}= f(\left | \alpha\right |)C_{\textnormal{L}}^{1}\sin(\left | \alpha \right |)+ (1-f(\left | \alpha\right |))C_{\textnormal{L}}^{2}\sin(2\left | \alpha \right |)
\end{equation}

\begin{equation}\label{eq:CD}
        C_{\textnormal{D}}= f(\left | \alpha\right |)(C_{\textnormal{D}}^{0}+C_{\textnormal{D}}^{1}\sin^2(\left | \alpha\right |)) + (1-f(\left | \alpha\right |))C_{\textnormal{D}}^{\pi/2}\sin^2(\left | \alpha\right |)
\end{equation}

\begin{equation}\label{eq:LCP}
        \ell_{\textnormal{CP}}/\ell = f(\left | \alpha\right |)(C_{\textnormal{CP}}^{0}-C_{\textnormal{CP}}^{1}\alpha^2) + C_{\textnormal{CP}}^{2}[1-f(\left | \alpha\right |)](1-\left | \alpha\right |/(\pi /2))
\end{equation}

\begin{equation}\label{eq:tran_l}
        L_{\textnormal{T}} = \frac{1}{2}\rho_f\ell C_{\textnormal{L}}\sqrt{{x_1}^2+(x_2- x_3 y_1\ell)^2}\left ( x_2- x_3y_1\ell, {x_1} \right )
\end{equation}

\begin{equation}\label{eq:rot_l}
        L_{\textnormal{R}} = -\frac{1}{2}\rho_f\ell^2 C_{\textnormal{R}}x_3\left ( x_2- x_3y_1\ell, {x_1} \right )
\end{equation}

\begin{equation}\label{eq:drag}
        D = -\frac{1}{2}\rho_f\ell C_{\textnormal{D}} \sqrt{{x_1}^2+(x_2- x_3 y_1\ell)^2}\left ( x_1,x_2- x_3y_1\ell \right )
\end{equation}

\begin{equation}\label{eq:tran_t}
        \tau_\textnormal{T} = -\frac{1}{2}\rho_f\ell \sqrt{{x_1}^2+(x_2 - x_3 y_1\ell)}\left [ C_{\textnormal{L}}{x_1}+C_{\textnormal{D}}(x_2 - x_3 y_1\ell) \right ] \left ( \ell_{\textnormal{CP}}- \ell_{\textnormal{CM}} \right )
\end{equation}

\begin{equation}\label{eq:arr}
        \tau_\textnormal{R} = -\frac{1}{128}\rho_f\ell^4 C_{\textup{D}}^{\pi /2} x_3  \left| x_3 \right| \left [ \left ( 2y_1+1 \right )^4 \pm \left ( 2y_1+1 \right )^4\right ]
\end{equation}

\begin{equation}\label{eq:x1}
        m \dot{x_1}=\left(m+\pi \rho_f \ell^2 / 4\right) x_3 x_2-(\pi \rho_f \ell^2 / 4) x_3^2 \ell_{C M}+L_T^{x^{\prime}}+L_R^{x^{\prime}}+D^{x^{\prime}}-m^{\prime} g \sin x_4
\end{equation}

\begin{equation}\label{eq:x2}
        \left(m+\pi \rho_f \ell^2 / 4\right) \dot{x_2}=-m x_3 x_1+(\pi \rho_f \ell^2 / 4) \dot{x_3} \ell_{C M}+L_T^{y^{\prime}}+L_R^{y^{\prime}}+D^{y^{\prime}}-m^{\prime} g \cos x_4
\end{equation}

\begin{equation}\label{eq:x3}
        I \dot{x_3}=\tau_T+\tau_R
\end{equation}

\begin{equation}\label{eq:x4}
        \dot{x_4}=x_3
\end{equation}

\begin{equation}\label{eq:x5}
        \dot{x_5}=x_1 \cos x_4-x_2 \sin x_4
\end{equation}

\begin{equation}\label{eq:x6}
        \dot{x_6}=x_1 \sin x_4+x_2 \cos x_4
\end{equation}

%% file: appendix.tex
\subsection{Equations}\label{app:equations}
\clearpage
\begin{sidewaystable}
\small
\centering
\caption{{{2D quasi-steady equations for falling plates with a displaced centre of mass \cite{Li2022model}, with six system variables ($x,y,\theta,{v}_{x'},{v}_{y'},\omega$). When integrated over time, the force coefficients are derived from the angle of attack (5-7), the forces are calculated (8-12), followed by the equations of motion (13-18). For simplicity the plate is assumed to be infinitesimally thin, and to always have an angle of attack within the region $\alpha \in [-\pi /2, 0]$.}}\vspace{4mm}}
\begin{tabular}{llll}
\cline{2-4}
 & \textbf{Constant(s)} & \textbf{Definition(s)} & \textbf{Value(s)} \\ \cline{2-4} 
 & $\ell$, $m$ & Plate length [m] and mass [kg] & 0.07, 3.175e-04 \\
 & $\rho_f$ & Fluid Density [$\textnormal{kg}/ \textnormal{m}^{3}$] & 1.225 \\
 & $\alpha_0$, $\delta$ & Critical $\alpha$ at stall, stall transition smoothness [$\degree$] & 14, 6 \\
 & \tiny{$C_{L}^{1}, C_{L}^{2}, C_{D}^{0}, C_{D}^{1}, C_{D}^{\pi/2}, C_{CP}^{0}, C_{CP}^{1}, C_{CP}^{2}, C_{R}$} & System-specific aerodynamic coefficients & \begin{tabular}[c]{@{}l@{}}0.23857, 2.8529, 0.36893, \\ 5.1822, 0.80751, 0.10598, \\ 4.9368, 1.4996, 1.73\end{tabular} \\
 & $a, b$ & Elliptical semi axes [m] & 0.03375, 5e-04 \\ \hline
No. & \textbf{Variable} & \textbf{PDE expression} &  \\ \hline
1 & CoM displacement, $\ell_{\textup{CM}}/\ell$ or $e_x$ & Defined by controller, in the range $[0.181,0.193]$ &  \\
2 & Moment of Inertia, $I$ [kgm$^2$] & $I = (m(a^2 + b^2)/(\rho_f  \ell^4))+1/32+(\ell_{\textup{CM}}/\ell)^2$ &  \\
3 & Angle of attack, $\alpha$ (p15) & $\tan\alpha = (v_{y'} - \omega\ell_{\textup{CM}})/v_{x'}\approx v_{y'}/v_{x'}$ &  \\
4 & Selection function, $f(\alpha)$ (5.2) & \multicolumn{2}{l}{$f=(1-\tanh((\alpha-\alpha_0)/\delta ))/2$} \\
5 & Lift coefficient, $C_{\textnormal{L}}(\alpha)$ (5.1, 5.3) & \multicolumn{2}{l}{$-C_{\textnormal{L}}= f(\left | \alpha\right |)C_{\textnormal{L}}^{1}\sin(\left | \alpha \right |)+ (1-f(\left | \alpha\right |))C_{\textnormal{L}}^{2}\sin(2\left | \alpha \right |)$} \\
6 & Drag coefficient, $C_{\textnormal{D}}(\alpha)$ (5.4, 5.5) & \multicolumn{2}{l}{$C_{\textnormal{D}}= f(\left | \alpha\right |)(C_{\textnormal{D}}^{0}+C_{\textnormal{D}}^{1}\sin^2(\left | \alpha\right |)) + (1-f(\left | \alpha\right |))C_{\textnormal{D}}^{\pi/2}\sin^2(\left | \alpha\right |)$} \\
7 & Center of pressure, $\ell_{\textnormal{CP}}(\alpha)$ (5.6, 5.7) & \multicolumn{2}{l}{$\ell_{\textnormal{CP}}/\ell = f(\left | \alpha\right |)(C_{\textnormal{CP}}^{0}-C_{\textnormal{CP}}^{1}\alpha^2) + C_{\textnormal{CP}}^{2}[1-f(\left | \alpha\right |)](1-\left | \alpha\right |/(\pi /2))$} \\
8 & Translational lift force, $L_{\textnormal{T}}$ (4.10) & \multicolumn{2}{l}{$L_{\textnormal{T}} = \frac{1}{2}\rho_f\ell C_{\textnormal{L}}\sqrt{{v_{x'}}^2+(v_{y'}- \omega \ell_{\textup{CM}})^2}\left ( v_{y'}- \omega\ell_{\textup{CM}}, {v_{x'}} \right )$} \\
9 & Rotational lift force, $L_{\textnormal{R}}$ (4.11) & \multicolumn{2}{l}{$L_{\textnormal{R}} = -\frac{1}{2}\rho_f\ell^2 C_{\textnormal{R}}\omega\left ( v_{y'}- \omega\ell_{\textup{CM}}, {v_{x'}} \right )$} \\
10 & Drag force, $D$ (4.13) & \multicolumn{2}{l}{$D = -\frac{1}{2}\rho_f\ell C_{\textnormal{D}} \sqrt{{v_{x'}}^2+(v_{y'}- \omega \ell_{\textup{CM}})^2}\left ( v_{x'},v_{y'}- \omega\ell_{\textup{CM}} \right )$} \\
11 & Torque from transl. forces, $\tau_\textnormal{T}$ (4.14) & \multicolumn{2}{l}{$\tau_\textnormal{T} = -\frac{1}{2}\rho_f\ell \sqrt{{v_{x'}}^2+(v_{y'} - \omega \ell_{\textup{CM}})}\left [ C_{\textnormal{L}}{v_{x'}}+C_{\textnormal{D}}(v_{y'} - \omega \ell_{\textup{CM}}) \right ] \left ( \ell_{\textnormal{CP}}- \ell_{\textnormal{CM}} \right )$} \\
12 & Aerodynamic rot. resistance, $\tau_\textnormal{R}$ (4.15) & \multicolumn{2}{l}{$\tau_\textnormal{R} = -\frac{1}{128}\rho_f\ell^4 C_{\textup{D}}^{\pi /2} \omega  \left| \omega \right| \left [ \left ( \frac{2\ell_{\textup{CM}}}{\ell}+1 \right )^4 \pm \left ( \frac{2\ell_{\textup{CM}}}{\ell}-1 \right )^4\right ]$} \\ \hline
\textbf{13} & Fixed frame $x$ velocity, $\dot{x}$ (4.4) & $\dot{x}=v_{x^{\prime}} \cos \theta-v_{y^{\prime}} \sin \theta$ &  \\
\textbf{14} & Fixed frame $y$ velocity, $\dot{y}$ (4.5) & \multicolumn{2}{l}{$\dot{y}=v_{x^{\prime}} \sin \theta+v_{y^{\prime}} \cos \theta$} \\
\textbf{15} & Angular velocity, $\dot{\theta}$ (4.6) & \multicolumn{2}{l}{$\dot{\theta}=\omega$} \\
\textbf{16} & Platewise $x'$ acceleration, $\dot{v}_{x'}$ (4.7) & \multicolumn{2}{l}{$m \dot{v_{x^{\prime}}}=\left(m+\pi \rho_f \ell^2 / 4\right) \omega v_{y^{\prime}}-(\pi \rho_f \ell^2 / 4) \omega^2 \ell_{C M}+L_T^{x^{\prime}}+L_R^{x^{\prime}}+D^{x^{\prime}}-m^{\prime} g \sin \theta$} \\
\textbf{17} & Platewise $y'$ acceleration, $\dot{v}_{y'}$ (4.8) & \multicolumn{2}{l}{$\left(m+\pi \rho_f \ell^2 / 4\right) \dot{v_{y^{\prime}}}=-m \omega v_{x^{\prime}}+(\pi \rho_f \ell^2 / 4) \dot{\omega} \ell_{C M}+L_T^{y^{\prime}}+L_R^{y^{\prime}}+D_{y^{\prime}}-m^{\prime} g \cos \theta$} \\
\textbf{18} & Angular acceleration, $\dot{\omega}$ (4.9) & \multicolumn{2}{l}{$I \dot{\omega}=\tau_T+\tau_R$} \\ \hline
\end{tabular}

\end{sidewaystable}